\documentclass[letterpaper]{article} % DO NOT CHANGE THIS
\usepackage{aaai25}  % DO NOT CHANGE THIS
\usepackage{times}  % DO NOT CHANGE THIS
\usepackage{helvet}  % DO NOT CHANGE THIS
\usepackage{courier}  % DO NOT CHANGE THIS
\usepackage[hyphens]{url}  % DO NOT CHANGE THIS
\usepackage{graphicx} % DO NOT CHANGE THIS
\usepackage{natbib}  % DO NOT CHANGE THIS AND DO NOT ADD ANY OPTIONS TO IT
\usepackage{caption} % DO NOT CHANGE THIS AND DO NOT ADD ANY OPTIONS TO IT
\usepackage{algorithm}
\usepackage{algorithmic}
\usepackage{amsmath,amssymb,amsfonts}
\usepackage{xcolor}
\usepackage{booktabs}
\usepackage{multirow}
\usepackage[capitalize]{cleveref}
\crefname{section}{Sec.}{Secs.}
\Crefname{section}{Section}{Sections}
\Crefname{table}{Table}{Tables}
\crefname{table}{Tab.}{Tabs.}
\crefname{figure}{Fig.}{Figs.}
\usepackage{newfloat}
\usepackage{listings}
\DeclareCaptionStyle{ruled}{labelfont=normalfont,labelsep=colon,strut=off} % DO NOT CHANGE THIS
\floatstyle{ruled}
\newfloat{listing}{tb}{lst}{}
\floatname{listing}{Listing}
\title{MambaLCT: Boosting Tracking via Long-term Context State Space Model}
\author{
    Xiaohai Li\textsuperscript{\rm 1,2}, Bineng Zhong\textsuperscript{\rm 1}\thanks{Corresponding author.}, Qihua Liang\textsuperscript{\rm 1}, Guorong Li\textsuperscript{\rm 3}, Zhiyi Mo\textsuperscript{\rm 2}\thanks{Corresponding author.}, Shuxiang Song\textsuperscript{\rm 1}
}
\affiliations{
    \textsuperscript{\rm 1}Key Laboratory of Education Blockchain and Intelligent Technology, Ministry of Education,\\ Guangxi Normal University, Guilin 541004, China\\
    \textsuperscript{\rm 2}Guangxi Colleges and Universities Key Laboratory of Intelligent Software, Wuzhou University, Wuzhou 543002, China\\
    \textsuperscript{\rm 3}Key Laboratory of Big Data Mining and Knowledge Management,\\ University of Chinese Academy of Sciences, Beijing 100101, China\\

bruc\_0619@stu.gxnu.edu.cn, bnzhong@gxnu.edu.cn, qhliang@gxnu.edu.cn\\
liguorong@ucas.edu.cn, zhiyim@gxuwz.edu.cn, songshuxiang@mailbox.gxnu.edu.cn\\

}
\usepackage{bibentry}
\begin{document}

\maketitle

\begin{abstract}
Effectively constructing context information with long-term dependencies from video sequences is crucial for object tracking. However, the context length constructed by existing work is limited, only considering object information from adjacent frames or video clips, leading to insufficient utilization of contextual information. To address this issue, we propose MambaLCT, which constructs and utilizes target variation cues from the first frame to the current frame for robust tracking. 
First, a novel unidirectional Context Mamba module is designed to scan frame features along the temporal dimension, gathering target change cues throughout the entire sequence. Specifically, target-related information in frame features is compressed into a hidden state space through selective scanning mechanism. The target information across the entire video is continuously aggregated into target variation cues. 
Next, we inject the target change cues into the attention mechanism, providing temporal information for modeling the relationship between the template and search frames.
The advantage of MambaLCT is its ability to continuously extend the length of the context, capturing complete target change cues, which enhances the stability and robustness of the tracker.
Extensive experiments show that long-term context information enhances the model's ability to perceive targets in complex scenarios. MambaLCT achieves new SOTA performance on six benchmarks while maintaining real-time running speeds.  Code and models are available at \url{https://github.com/GXNU-ZhongLab/MambaLCT}.
\end{abstract}

\section{Introduction}

Visual tracking involves locating the object in subsequent frames based on the given initial template information. In long and complex sequences, the target's appearance and motion state may change. However, traditional tracking algorithms \cite{siamban,hu2023transformer,ye2023positive} rely solely on the initial frame to locate the target in subsequent frames. They lack an understanding about change in objects through the sequence, leading to reduced stability and robustness of the tracker.
\begin{figure}[t]
      \centering
       \includegraphics[width=1\linewidth]{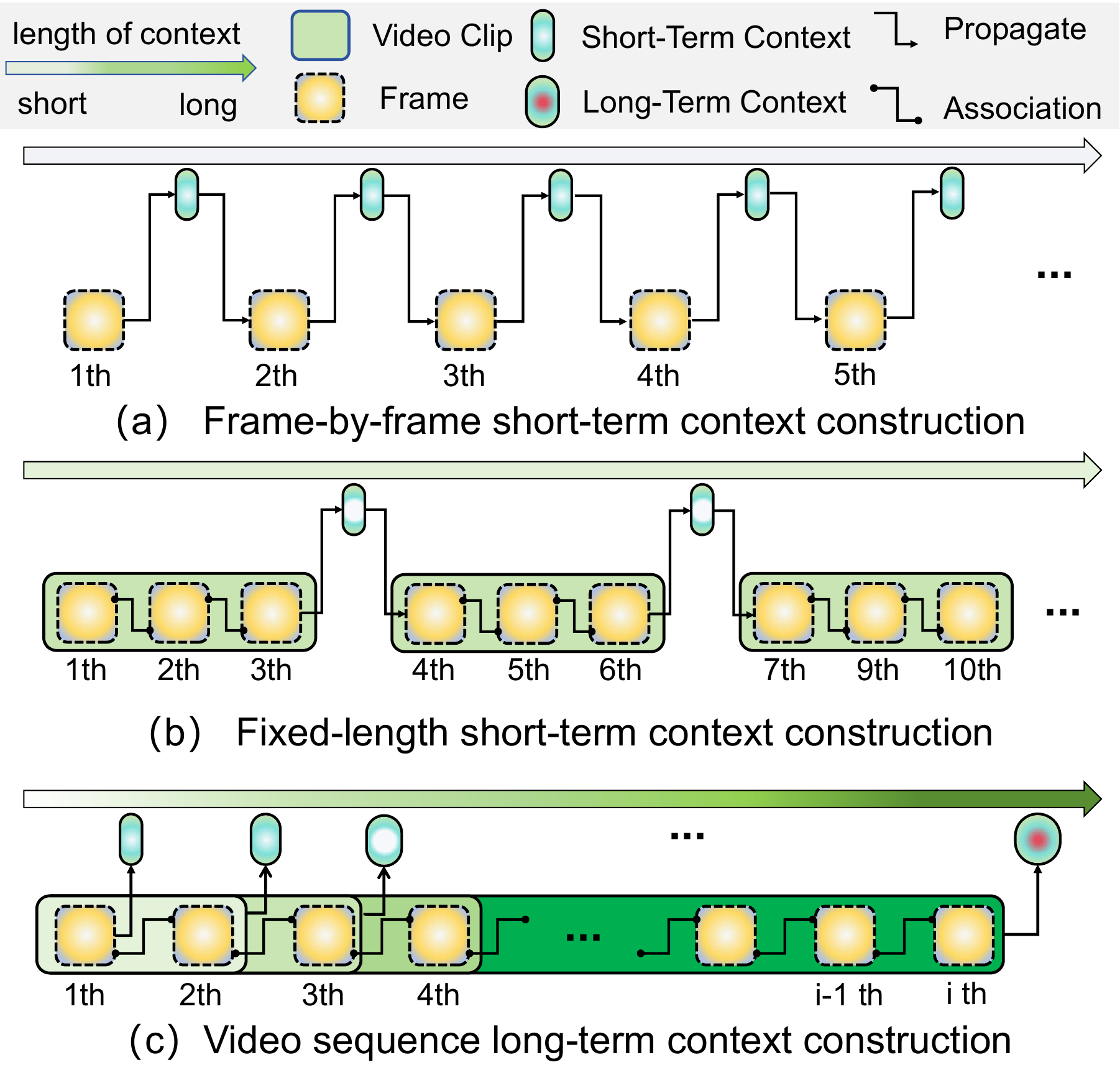}

       \caption{Comparison between current SOT context information construction paradigm and our method. (a) and (b) construct short-term context information within the scope of frames or video clips \cite{our3EVPTrack,our1ODtrack}. (c) Our propose MambaLCT analyzes the complete video sequence to construct long-term context information. }

       \label{figure:motivation}
       % \vskip -0.1in
\end{figure}

Recently, an increasing number of works have focused on the context information of a target in video sequences. Context information records the changes in the target's position, scale, appearance, and other trends. Effectively constructing context information from video sequences plays a crucial role in tracking algorithms. The length of the contextual information constructed in the current work is limited, which directly affects the range within which the tracker can perceive target changes.
The current methods for constructing short-term context information can be summarized as follows: (1) \textit{The frame-by-frame short-term context construction method}, as shown in \cref{figure:motivation}(a). KeepTrack \cite{Martin} and EVPTrack\cite{our3EVPTrack} explores the associations between consecutive frames to enhance target perception. This association is a fragmented context, which focuses on target changes in adjacent frames. (2) \textit{The fixed-length context construction method}, as shown in \cref{figure:motivation}(b), models the temporal sequence of images within a fixed length. ODTrack \cite{our1ODtrack} and AQATrack \cite{our2AQATrack} capture target context information using video clips and sliding windows, length of 4. 
The methods mentioned above rely on the Transformer for learning spatio-temporal information. While the Transformer performs exceptionally well in learning appearance features, its quadratic computational complexity restricts the length of context information it can handle. Recently, Mamba has shown great potential for computational efficiency in building long-term dependencies. Nevertheless, the performance of Mamba for non-autoregressive appearance feature learning is often underwhelming. 
Mamba and Transformer complement each other in learning appearance and context information, which has led us to consider: \textit{Can our method combine the strengths of both to extend the context length, thereby enabling the learning of more robust appearance features?} 

In this paper, we introduce a new tracking framework, namely MambaLCT, which aims to fully utilize target contextual information, as demonstrated in \cref{figure:motivation}(c). MambaLCT gradually expands the range of context information to cover from the initial frame to the current frame.
Specifically, we use Transformer in an autoregressive manner to learn the appearance features of the search image. Then, the autoregressive appearance features are continuously fed into the Context Mamba module during scanning frame features. Through Mamba's selective scanning mechanism, the information relevant to the target in the appearance features of historical search frames is continuously aggregated into the final state space via the transfer of hidden states. Compared to fragmented context information, unified modeling of all historical appearance features ensures the consistency and completeness of context information. Finally, the context information from the first frame to the current frame provides all historical target change information for modeling the appearance features of the next frame, thereby enhancing the accuracy and robustness of the tracker. The main contributions of this work are as follows:

\begin{itemize}
    \item We propose a tracker named MambaLCT, which can effectively capture the long-term behavior and overall motion of the target, providing more comprehensive contextual information.
    \item We design a Context Mamba module that can effectively and with low resource consumption construct long-term contextual information along the temporal dimension.
    \item Our method has achieved a new state-of-the-art tracking performance on six visual tracking benchmarks, including LaSOT, LaSOT$_{\rm{ext}}$, GOT-10K, TrackingNet, TNL2K and UAV123.
\end{itemize}

\section{Related Work}
\textbf{Tracking paradigm based on initial template.} The initial template is reliable target information manually annotated. Whether it is the earlier two-stream framework trackers or the current single-stream framework trackers, they all heavily rely on the features of the initial template frame to complete tracking. Therefore, most tracking algorithms follow the search-template image matching paradigm. For example, SiamFC \cite{siamfc} and SiamBAN \cite{siamban}, which are based on Siamese two-stream trackers, use networks with shared parameters to extract features from the search image and the template image separately. Then, they calculate the regions in the search image that are similar to the template image through cross-correlation operations. With the introduction of Transformers into the field of computer vision, an increasing number of works have started using Transformers for representation learning. An example of this is OSTrack \cite{ostrack}, which addresses the issue of poor target perception caused by the two-stream, two-stage tracking framework. Although the tracking paradigm relying on initial template features has achieved competitive performance, with the advent of long sequence tracking benchmarks \cite{LaSOT}, the object features recorded by the initial template are insufficient to describe the target in subsequent frames. In this paper, our aim is to construct a tracker that can perceive long-term target changes.
\begin{figure*}[t]
      \centering
       \includegraphics[width=0.9\linewidth]{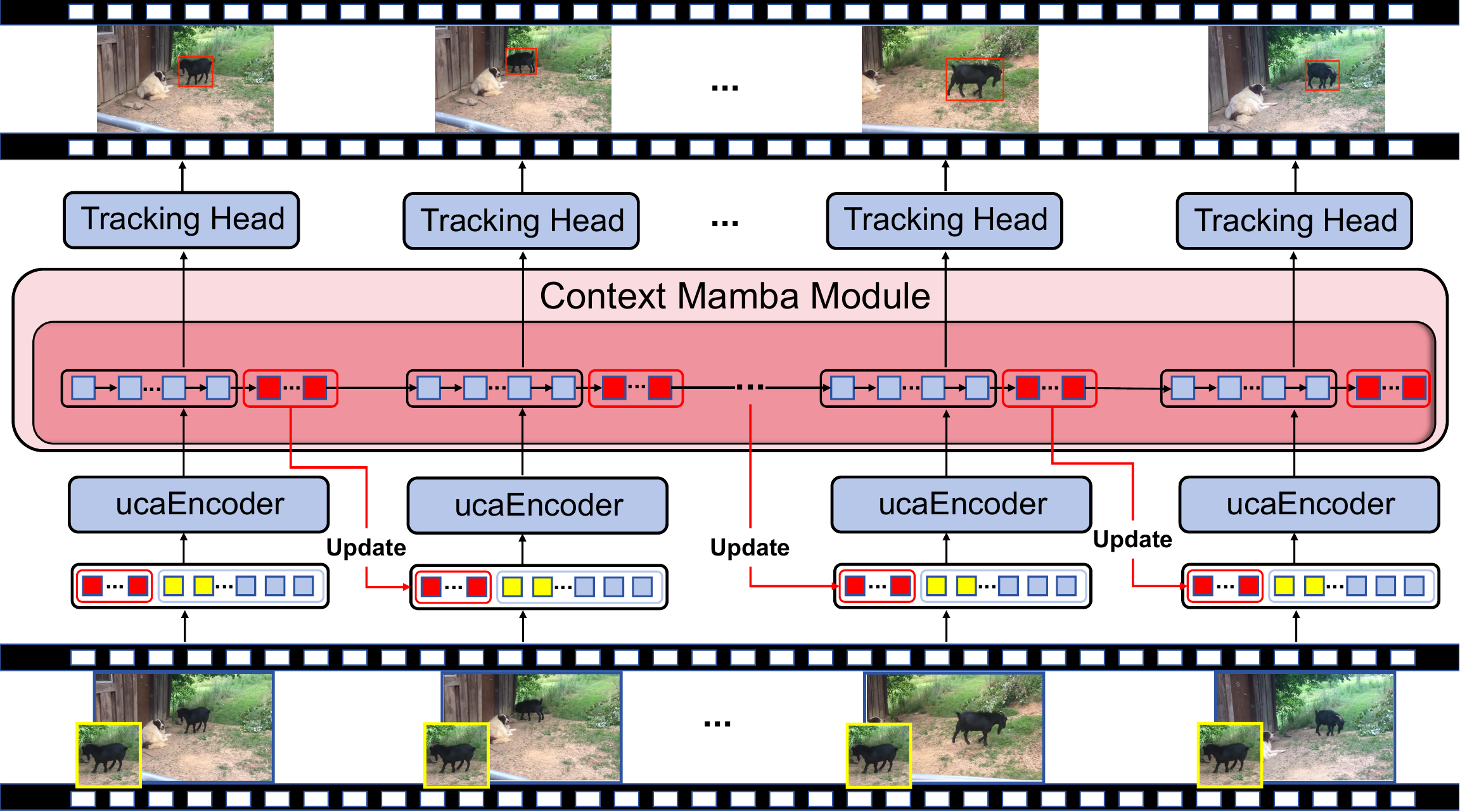}
    
       \caption{Overview of our framework. The input video frames are converted into tokens through patch embedding. Then, these tokens, along with the contextual information, are fed into the ucaEncoder for unified modeling of the contextual and appearance information. During the temporal scanning process, the representational information of the images is continuously fed into the Context Mamba module to construct the target's change cues.}
       \label{figure:framework}
       \vskip -0.1in
\end{figure*}

\textbf{Tracking paradigm based on context information.} Video sequences contain rich contextual information, recording the state of the target at multiple points in time. To overcome the limitations of the initial template, tracking frameworks that heavily utilize contextual information have begun to emerge. Such as VideoTrack \cite{VideoTrack1} proposed a method using a video Transformer to encode spatio-temporal information in videos. Some exciting works were proposed last year, utilizing frame-by-frame propagation methods to transmit cross-frame contextual information. ODTrack \cite{our1ODtrack} uses a token to transmit cross-frame information from the first frame to the last frame. The token learns cross-frame contextual information by being encoded together with image pairs. This method is simple and efficient, but it can lead to redundant computations. AQATrack \cite{our2AQATrack} designs a Transformer-based spatio-temporal information fusion module to extract information between frames and propagate it through queries. The context length constructed by these aforementioned methods can only focus on target changes in adjacent frames or within a window, resulting in limited help from the context for the tracker. To address this issue, we propose a method for constructing context that can focus on target changes in all historical search frames.

\textbf{Mamba for computer vision.} Mamba model has demonstrated excellent semantic modeling capabilities in the field of natural language processing. Some works \cite{vim,localmamba} have successfully applied the Mamba model to the field of computer vision. Compared to the quadratic complexity of Transformers, Mamba has great potential in handling long sequences of data with its linear complexity. Vim \cite{vim} designed a bidirectional state space to enhance Mamba’s sensitivity to visual data and meet the contextual requirements of visual understanding. \cite{videomamba} proposed VideoMamba, which introduces Mamba into the video domain, addressing issues of local redundancy and global dependencies in video understanding. Mamba4D \cite{mamba4d} designed an efficient 4D point cloud video backbone framework to unify the irregular and unordered distribution of points.
Due to the inherent mechanism of SSM, Mamba has autoregressive properties. However, appearance features in tracking tasks do not conform to this characteristic. In this work, we separate the modeling of temporal and spatial information in tracking tasks, applying Mamba to the temporal modeling.

\section{Method}
In this section, we introduce how MambaLCT constructs and utilizes long-term contextual. First, we briefly present our framework, followed by an explanation of how we construct contextual information along the temporal dimension and unify the modeling of appearance and context information. Finally, we describe the training pipeline.

\subsection{Overview}
An overview of the MambaLCT pipeline is shown in \cref{figure:framework}. This framework is very straightforward, consisting of three main components: the ucaEncoder, the Context Mamba module, and the tracking head. 
First, the initial template \({t}\in\mathbb{R}^{3\times H_t\times W_t}\) and the video frames \({s}\in\mathbb{R}^{3\times H_s\times W_s}\) are input into the pipeline sequentially, \(H\) and \(W\) represent the height and width of the image. \({s}\) and \({t}\) are converted into 1D tokens \({s}_p\in\mathbb{R}^{N_s\times D}\) and \({t}_p\in\mathbb{R}^{N_t\times D}\) through the image patch embedding process. Here \(N_{s}=H_{s}W_{s}/16^{2}\), \(N_{t}=H_{t}W_{t}/16^{2}\), \(D=512\). These tokens are then concatenated with the contextual information tokens \({c}_p\in\mathbb{R}^{N_c\times D}\) and fed into the ucaEncoder, as shown in \cref{figure:Mamba}, \(N_c\) is the length of \({c}_p\). Then, we design a Context Mamba Module to construct long-term contextual information. The search frame feature flow output by the ucaEncoder is continuously fed into the Context Mamba module, where the target information with long-term dependencies is aggregated into the \({c}_p\).
Finally, the search image features enter the tracking head for classification and regression to obtain the results.

\subsection{Preliminaries}
Mamba excels at handling sequences with long-term dependencies.
The Mamba model is inspired by State Space Models. SSMs are a class of models used for handling time series data, typically mapping an input sequence \(x(t)\in\mathbb{R}\) an output sequence \(y(t)\in\mathbb{R}\). The mapping process is as follows:
\begin{equation}
\label{eq1}
\begin{aligned}h'(t)&=\mathbf{A}h(t)+\mathbf{B}x(t),\\y(t)&=\mathbf{C}h(t),\end{aligned}
\end{equation}
where \(h(t)\) is the hidden state, \(\mathbf{A}\) is the evolution parameter, and \(\mathbf{B}\), \(\mathbf{C}\) are the projection parameters.

To effectively apply the Mamba model in the field of deep learning algorithms, we need to discretize the continuous parameters \(\mathbf{A}\) and \(\mathbf{B}\). A common discretization method is zero-order hold (ZOH), which is defined as follows:
\begin{equation}
\label{eq2}
\begin{aligned}&\overline{\mathbf{A}}=\exp{(\boldsymbol{\Delta}\mathbf{A})},\\&\overline{\mathbf{B}}=(\boldsymbol{\Delta}\mathbf{A})^{-1}(\exp{(\boldsymbol{\Delta}\mathbf{A})}-\mathbf{I})\cdot\boldsymbol{\Delta}\mathbf{B},\end{aligned}\end{equation}
where \(\Delta\) is a timescale parameter for discretization. The discretized version of \cref{eq1} and \cref{eq2} can be rewritten as:
\begin{equation}\begin{aligned}&h_{t}=\overline{\mathbf{A}}h_{t-1}+\overline{\mathbf{B}}x_{t},\\&y_{t}=\mathbf{C}h_t.\end{aligned}\end{equation}

\subsection{Context Mamba Module}
\begin{figure}[t]
      \centering
       \includegraphics[width=1\linewidth]{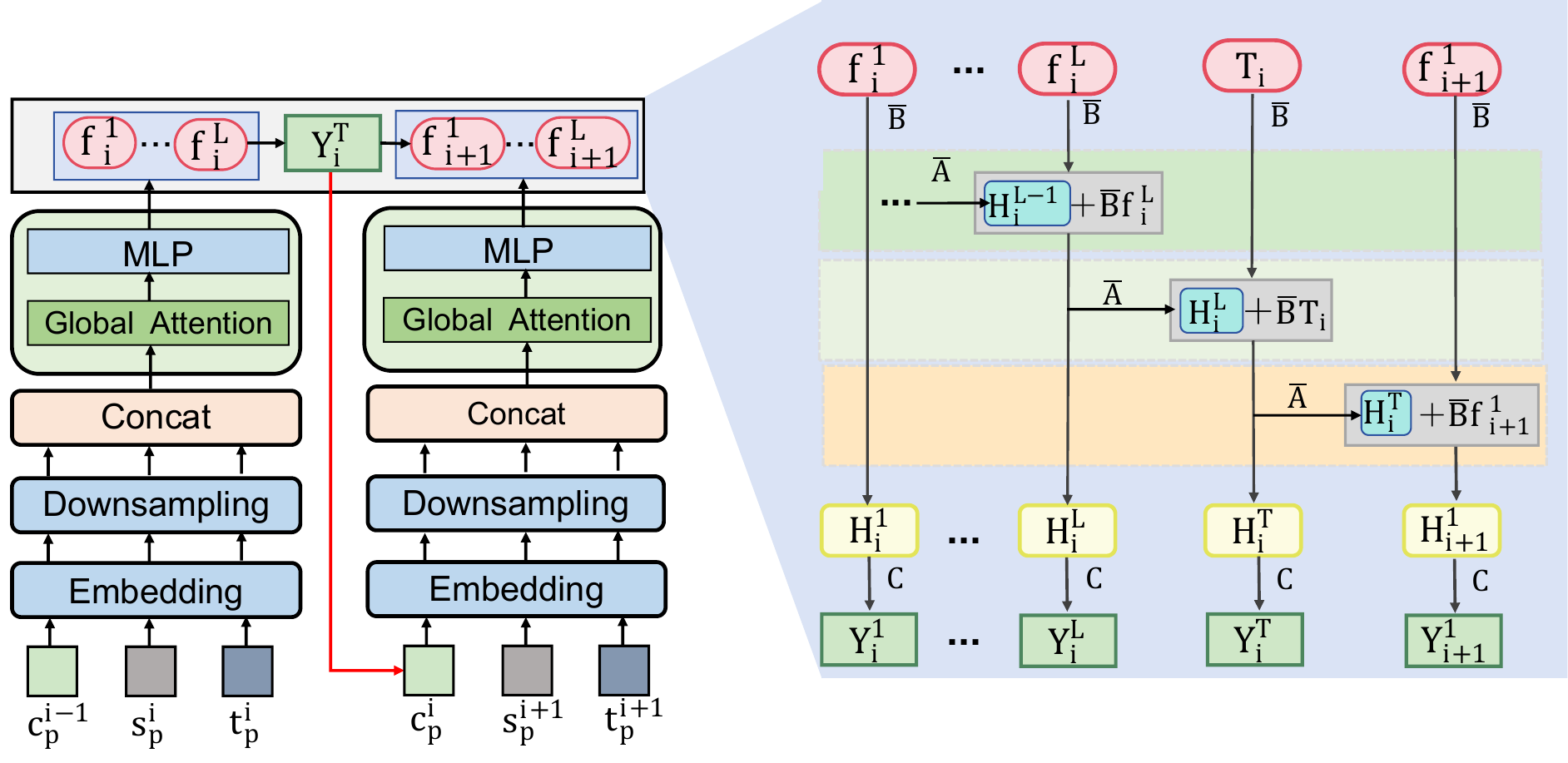}
    
       \caption{Illustration of the process of constructing and propagating context information. On the left is the structure of the ucaEncoder, and on the right is the process of constructing contextual information.}
       \label{figure:Mamba}
       % \vskip -0.1in
\end{figure}
Existing works primarily extract contextual information from adjacent frames or images within a nearby window.
\cite{our1ODtrack,our2AQATrack} proposing a novel method to construct context information. They increase the context length by focusing on target changes within a fixed window. Their approach to constructing contextual information can be summarized as follows:
\begin{equation}
\label{eq5}
T\leftarrow\ f:\{T_{i-l},...T_{i},S_{i-l},...,S_{i}\},
\end{equation}
where the commonly used \(f\) is generally based on Transformer methods. \(S\) represents the search image, \(T\) denoted the template image. \(l\) is the length of context information.

To fully utilize the contextual information in videos, we propose a Mamba-based context construction method, which can be described as follows:
\begin{equation}
 \label{eq4}
T\leftarrow\ Mamba:\{S_1,S_2,S_3...,S_i\},
\end{equation}

stick to Mamba's strengths, we do not use Mamba to learn the spatial features of objects. Instead, we directly perform sequential modeling on the historical search frame flow. From \cref{eq4} and \cref{eq5}, we can see that short-term contextual information is limited to the target information of the adjacent \(l\) frames, whereas our method can capture target information from the first frame to the current frame.

As shown in \cref{figure:Mamba}, the search features from the first frame to the \(i\) frame are continuously fed into the Context Mamba Module, as follows:
\begin{equation}
input=[f_1^1W;\cdots;f_1^LW;\cdots;f_i^1{W};\cdots;{f}_i^L{W}],
\end{equation}
where \({f}_i^j\) represents the \(j\)-th token in the feature of the \(i\)-th frame, \({W}\) is the learnable linear projection matrix, \(L\) is the length of image token. 
Through the selective scanning mechanism, the information related to the target in the \(f_i\) is compressed into the \(H_i^L\). This process is described as:
\begin{equation}
\begin{aligned}
&H_i^t=\overline{\mathbf{A}}H_i^{t-1}+\overline{\mathbf{B}}f_i^t, &t\leq L
\end{aligned}
\end{equation}
where \(H_i^t\) represents the current aggregated hidden state. Information about the target from frames 1 to \(i\) in the video sequence is also continuously aggregated into \(Y_i^T\) through the hidden state, Context information is transmitted from \(f_i\) to \(f_{i+1}\) through \(Y_i^T\), as follows:
\begin{equation}
\begin{aligned}
&H_i^{T}=\overline{\mathbf{A}}H_i^{L}+\overline{\mathbf{B}}T_{i},
\\&H_{i+1}^{1}=\overline{\mathbf{A}}H_i^{T}+\overline{\mathbf{B}}f_{i+1}^1,
\\&Y_i^{T}=\mathbf{C}H_i^T,
\end{aligned}
\end{equation}
where \(T_{i}\) is an empty token used to record all historical target information from the past \(i\) frames, and \(H_i^T\) serves as the medium for information transmission between frames. \(Y_i^{T}\) is used to update the \(c_p\).

\subsection{Unified Context and Appearance Modeling}
The ucaEncoder is a critical component of our framework, capable of achieving unified modeling of contextual and appearance information. The structure of this encoder is shown in \cref{figure:Mamba}, and it is implemented based on a transformer architecture. Unlike the vanilla ViT, which directly downsamples the input image with a stride of 16, we apply a hierarchical ViT to perform multiple stages of downsampling to avoid potential information loss caused by such a large stride. 
Specifically, we inject \(c_p\) into the attention operations of \(S_p\) and \(t_p\). Before this, we first perform two stages of downsampling on these three inputs, followed by global attention operations. This process can be summarized as follows:
\begin{equation}\begin{aligned}
&i_c,i_s,i_t=Ds(c_p,s_p,t_p),
\\&f_{input}=Concat(i_c,i_s,i_t),
\\&f_{i}=MLP(attention(f_{input})),
\end{aligned}\end{equation}
where \(Ds\) denotes the downsampling operation, \(f_i\) represents the features of the current frame. \(Y_i^{T}\), which contains all target change information from the first \(i\) frames, introduces contextual information into the relationship modeling between the template and search frames by updating \(c_p\). This effectively enhances the target change cues, thereby improving the tracker’s perception of the target.

\subsection{Training and Loss Function}
Common tracking dataset sampling methods involve randomly selecting video frames from a random dataset. These random images may come from different video sequences, making it impossible to learn context information. Our sampling method involves selecting video clips of a certain length from sequences within a random dataset, ensuring that there is contextual connection between the images. The context tokens from the previous frame, search frame patches, and template frame patches are concatenated and input into the ucaEncoder for interaction. Finally, the image features are fed into the tracking head for bounding box prediction. The tracking head consists of classification and regression head. We use classification loss, regression loss, and GIoU loss as training constraints. The total loss can be formulated as:
\begin{equation}L=L_{cls}+\lambda_{1}L_{1}+\lambda_{2}L_{GIoU},\end{equation}
where \(\lambda_{1}\) and \(\lambda_{2}\) are two manually set loss weights.

\section{Experiments}
\subsection{Implementation Details}
\textbf{Experimental environment.} Our tracker implementation is based on Python 3.8 and Pytorch 1.13.1. Training and testing were conducted on two NVIDIA A100 GPUs. The tracking speed test was performed on a Tesla V100. 
\begin{table}[h]
    \centering
    
    \fontsize{9}{10.8}\selectfont
    \resizebox{0.48\textwidth}{!}
    {
    \begin{tabular}{c|cccccc}
     \toprule
     \multicolumn{1}{c|}{Tracker} & \multicolumn{1}{c}{Type} & \multicolumn{1}{c}{Resolution}
     & \multicolumn{1}{c}{Params} & \multicolumn{1}{c}{MACs}& \multicolumn{1}{c}{Speed}& \multicolumn{1}{c}{Device} \\
     \midrule%第二道横线
     SeqTrack-384B & ViT-B & \(384\times384\) & 89M &148G & 12.8fps & Tesla V100 \\
     AQATrack-256 & HiViT-B & \(384\times384\) & 72M & 25.8G & 67.6fps & Tesla V100 \\
     AQATrack-384 & HiViT-B & \(384\times384\) & 72M & 58.3G & 44.2fps & Tesla V100 \\
     Our-256 & HiViT-B & \(256\times256\) & 72M &25G & 58.6fps & Tesla V100 \\
     Our-384 & HiViT-B & \(384\times384\) & 72M &58G & 45.3fps & Tesla V100 \\
     \bottomrule

    \end{tabular}
    }
    \caption{Comparison of model parameters, MACs(G), and Spees(fps). These test results were obtained on the same machine.}
    \label{tab:model}
\end{table}

\textbf{Model parameter.} To demonstrate the robustness of our tracker to different image scales, we designed two different trackers with varying image input resolutions, as shown in \cref{tab:model}.

\textbf{Training Details.} During the training process, we set the video clip sampling length to 2 and the sampling quantity to 30000. The datasets used for training are GOT10K \cite{got10k}, LaSOT \cite{LaSOT}, COCO \cite{coco}, and TrackingNet \cite{trackingnet}. We use HiViT-Base \cite{hivit} as the backbone network to extract visual features, with its parameters initialized using MAE \cite{mae}. The length of the cross-frame token is set to 1. The Mamba cross-frame information learning network uses Vim-Small \cite{vim} and is initialized with its checkpoint. We made some modifications to Vim, employing a unidirectional scanning method. During training, MambaLCT uses the AdamW \cite{adamW} optimizer to adjust model parameters. The learning rate for the backbone network is set to \(2\times10^{-4}\), while other parameters have a learning rate ten times higher. We train the model for a total of 300 epochs, and learning rate decay begins at the 240th epoch, with a decay rate set to \(1\times10^{-4}\). The same learning rate and decay rate are applied when training on the GOT10K dataset, but the model is trained for only 150 epochs, with the decay starting at the 120th epoch. The batch size for training is set to 16. In the loss function, \(\lambda_{1}=5\) and \(\lambda_{2}=2\).
\begin{figure}
    \centering
    \includegraphics[width=0.45\textwidth]{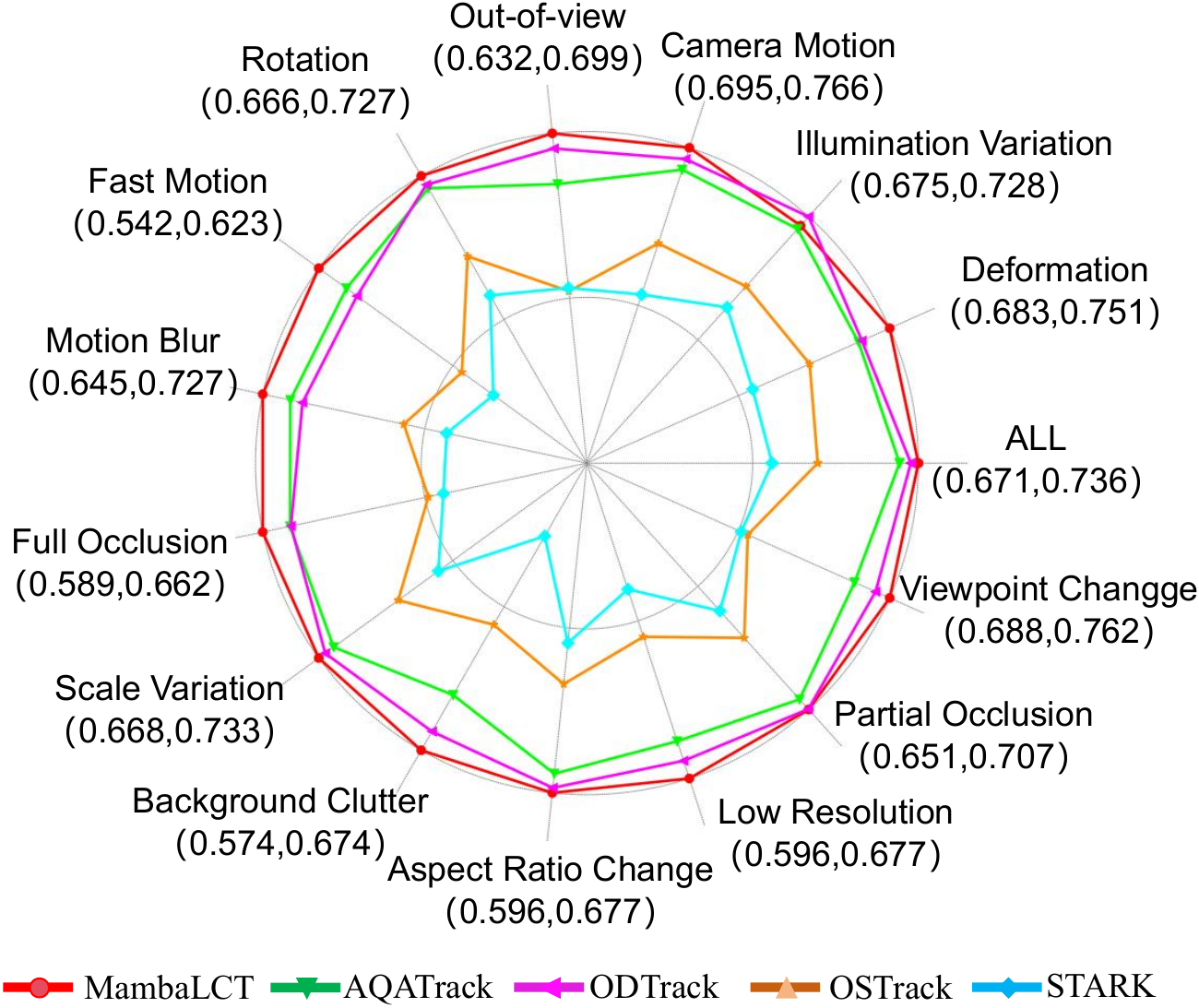}
    \caption{Attribute-based evaluation on the LaSOT test set. AUC score is used to rank different trackers.}
    \label{fig:eao_lasot}
\end{figure}
% \begin{figure}[t]
%       \centering
%        \includegraphics[width=1\linewidth]{picture/eao.pdf}
    
%        \caption{Visualization of the attention to the search frame by the cross-frame context information. As the length of the context information increases, the attention to the target becomes more focused. }
%        \label{figure:att}
%        % \vskip -0.1in
% \end{figure}

\textbf{Inference Details.} Unlike most tracking methods, in our frame-by-frame testing process, we use the search frame features and cross-frame tokens to construct a cross-frame information flow that includes the entire test sequence's image information. The cross-frame tokens act as a medium for information transmission between frames. Thanks to the Mamba model's advantages in handling long sequence data, our tracker achieves excellent speed and performance, as shown in \cref{tab:model}.

\begin{table*}[t]
% \caption{ Performance comparisons with state-of-the-art trackers on the test set of LaSOT, $\rm LaSOT_{ext}$, GOT-10K, and TrackingNet. We add a symbol * over GOT-10k to indicate that the corresponding models are only trained with the GOT-10k training set. The top two results are highlighted with {red} and {\color{blue}blue} fonts, respectively.}
\centering
\resizebox{\textwidth}{!}{
    \fontsize{9}{10.8}\selectfont
    \begin{tabular}{l|c|ccc|ccc|ccc|ccc}
    \toprule
    \multicolumn{1}{c|}{\multirow{2}{*}{Method} }
    & \multicolumn{1}{c|}{\multirow{2}{*}{Source}} 
    & \multicolumn{3}{c|}{LaSOT} 
    & \multicolumn{3}{c|}{$\rm LaSOT_{ext}$} 
    & \multicolumn{3}{c|}{GOT-10K$^*$} 
    & \multicolumn{3}{c}{TrackingNet}\\ 
    \cline{3-14}
    && AUC & $\rm P_{norm}$ & P    & AUC & $\rm P_{norm}$ & P     & AO & $\rm SR_{0.5}$ & $\rm SR_{0.75}$     & AUC & $\rm P_{norm}$ & P  \\
    \midrule
    MambaLCT-256 & our & \textbf{{71.8}} & \textbf{{83.0}} & \textbf{{79.4}}    & \textbf{{51.6}} & \textbf{{64.0}} & \textbf{{59.0}}    & \textbf{{74.8}} & \textbf{{85.4}} & \textbf{{72.1}}    &\textbf{{84.3}}   &\textbf{{89.2}}  &\textbf{{83.9}}\\
    \midrule
     AQATrack-256\cite{our2AQATrack}&CVPR24  & \underline{71.4} & \underline{81.9} & \underline{78.6}    & \underline{51.2} & \underline{62.2} & \underline{58.9}& \underline{73.8} & \underline{83.2} & \underline{72.1}    & 83.8 & 88.6 & 83.1 \\
    F-BDMTrack-256\cite{F-BDMTrack}&ICCV23  & 69.9 & 79.4 & 75.8    & 47.9 & 57.9 & 54.0    & 72.7 & 82.0 & 69.9    & 83.7 & 88.3 & 82.6 \\
    ROMTrack-256\cite{ROMTrack} & ICCV23    & 69.3 & 78.8 & 75.6    & 48.9 & 59.3 & 55.0    & 72.9 & 82.9 & 70.2    & 83.6 & 88.4 & 82.7 \\
    ARTrack-256\cite{mT1ARTrack} & CVPR23      & {70.4} & 79.5 & {76.6}  & 46.4 & 56.5 & 52.3    & 73.5 & 82.2 & 70.9    & \underline{84.2} & \underline{88.7} & \underline{83.5} \\
    GRM\cite{grm} &CVPR2023       & 69.9 & 79.3 & 75.8    & - & - & -    & 73.4 & 82.9 & 70.4    & 84.0 & 88.7 & 83.3\\
    OSTrack-256\cite{ostrack} &ECCV22       & 69.1 & 78.7 & 75.2    & 47.4 & 57.3 & 53.3    & 71.0 & 80.4 & 68.2    & 83.1 & 87.8 & 82.0\\
    VideoTrack\cite{VideoTrack1}  & CVPR23   & 70.2 & - & 76.4       & -    & -    & -       & 72.9 & 81.9 & 69.8    & 83.8 & \underline{88.7} & 83.1 \\
    AiATrack-320\cite{aiatrack}&ECCV22          & 69.0 & 79.4 & 73.8    &-     &-     &-        & 69.6 & 80.0 & 63.2    & 82.7 & 87.8& {80.4} \\
    MixFormer-22k\cite{Mixformer}&CVPR22    & 69.2 & 78.7 & 74.7    & -    & -    & -       & 70.7 & 80.0 & 67.8    & 83.1 & 88.1 & 81.6 \\
    STARK\cite{upT1STARK} & ICCV21 & 67.1 & 77.0 & -       & -    &-     & -       & 68.8 & 78.1 & 64.1    & 81.3 & 86.1 &-     \\
    TransT \cite{TransT}& CVPR21 & 64.9 & 73.8 & 69.0    & -    & -    & -       & 67.1 & 76.8 & 60.9    & 81.4 & 86.7 & 80.3 \\
    Ocean \cite{Ocean}&  ECCV 20            & 56.0 & 65.1 & 56.6    &-     & -    &-        & 61.1 & 72.1 & 47.3    & -    & -   &-     \\
    SiamRPN++\cite{siamrpn++}&CVPR19        & 49.6 & 56.9 & 49.1    & 34.0 & 41.6 & 39.6    & 51.7 & 61.6 & 32.5    & 73.3 & 80.0 & 69.4\\
    ECO \cite{ECO} & ICCV 17                & 32.4 & 33.8 & 30.1    & 22.0 & 25.2 & 24.0    & 31.6 & 30.9 & 11.1    & 55.4 & 61.8 & 49.2 \\
    SiamFC \cite{siamfc} & ECCVW16          & 33.6 & 42.0 & 33.9    & 23.0 & 31.1 & 26.9    & 34.8 & 35.3 & 9.8     & 57.1 & 66.3 & 53.3 \\
    \midrule
    
    \multicolumn{14}{l}{\multirow{1}{*}{\textit{Some Trackers with Higher Resolution}} }\\
    \midrule
    OSTrack-384\cite{ostrack}&ECCV22        & 71.1 & 81.1 & 77.6    & 50.5 & 61.3 & 57.6    & 73.7 & 83.2 & 70.8    & 83.9 &88.5  &83.2  \\
    SeqTrack-B384\cite{seqtrack} & CVPR23   & 71.5 & 81.1 & 77.8    & 50.5 & 61.6 & 57.5    & 74.5 & 84.3 & 71.4    & 83.9 & 88.8 & 83.6 \\
    ROMTrack-384\cite{ROMTrack} & ICCV23    & 71.4 & 81.4 & 78.2    & 51.3 & 62.4 & 58.6  & 74.2 & 84.3 & 72.4    & 84.1 & 89.0 & 83.7 \\
    F-BDMTrack-384\cite{F-BDMTrack}&ICCV23  & 72.0 & 81.5 & 77.7    & 50.8 & 61.3 & 57.8    & 75.4 & 84.3 & 72.9    & 84.5 & 89.0 & 84.0\\
    ARTrack-384\cite{mT1ARTrack} & CVPR23 & 72.6& 81.7 & 79.1 & 51.9 & 62.0 & 58.5& 75.5& 84.3 & 74.3   & \underline{85.1} & 89.1  & 84.8 \\
    ODTrack\cite{our1ODtrack}&AAAI24  & \underline{73.2} & \underline{83.2} & \underline{80.6} & 52.4 & 63.9 & 60.1 & \textbf{{77.0}} & \textbf{{87.9}} & \textbf{{75.1}} & \underline{85.1} & \textbf{{90.1}} & \underline{84.9}\\
    AQATrack-384\cite{our2AQATrack}&CVPR24  & 72.7 & 82.9 & 80.2 &\underline{52.7} & \underline{64.2} & \underline{60.8} & 76.0 & 85.2 & \underline{74.9} & 84.8 & 89.3 & 84.3\\
    \midrule
    % MambaLCT-384 & Ours & \textbf{{73.6}} & \textbf{{84.1}} & \textbf{{81.6}}    & \textbf{{53.3}} & \textbf{{64.8}} & \textbf{{61.4}}    & \underline{76.2}} & \underline{86.7}} & {74.3}   & \textbf{{85.2}}  & \underline{89.8}}  &\textbf{{85.2}}\\
    MambaLCT-384 & Ours & \textbf{73.6} & \textbf{{84.1}} & \textbf{{81.6}}    & \textbf{{53.3}} & \textbf{{64.8}} & \textbf{{61.4}}    & \underline{76.2} & \underline{86.7} & {74.3}   & \textbf{{85.2}}  & \underline{89.8}  &\textbf{{85.2}}\\
    \bottomrule    
    \end{tabular}
    }
\caption{Comparison with state-of-the-arts on four popular benchmarks: LaSOT, $\rm LaSOT_{ext}$, GOT-10K, and TrackingNet. Where * denotes for trackers only trained on GOT10K. Best in bold, second best underlined.}
\label{tab:result}
\end{table*} 
\subsection{State-of-the-art Comparisons}
\textbf{LaSOT.} The LaSOT dataset contains 1400 high-quality video sequences with a total duration of over 3800 minutes. Each video sequence has an average length of approximately 2500 frames. The videos in this dataset encompass rich cross-frame information. With similar resource consumption, our model achieved new state-of-the-art performance at different image input resolutions. As shown in \cref{tab:model}, MambaLCT-256 achieved 71.8\%, 83.0\%, and 79.4\% in AUC, P$_{\rm{Norm}}$, and precision score, respectively, across the three evaluation metrics.  Compared to ODTrack-B \cite{our1ODtrack}, MambaLCT-384 achieved improvements of 0.4\%, 0.9\%, and 1.0\% in AUC, P$_{\rm{Norm}}$, and precision score, respectively.

\textbf{LaSOT$_{\rm{ext}}$.} LaSOT$_{\rm{ext}}$ is an extension supplement to the LaSOT dataset, encompassing 15 categories and 150 videos. Our proposed method for building long-term dependency cross-frame contextual information performs excellently in long video sequences. Our MambaLCT model achieved state-of-the-art performance with an input image resolution of 256, attaining 51.6\% AUC, 64.0\% P$_{\rm{Norm}}$, and 59\% precision score. MambaLCT-384 achieved an improvement of 0.6\% in all three evaluation metrics compared to AQATrack-384. Compared to trackers using the same backbone network and input resolution, this result demonstrates that our proposed method offers superior performance.

\textbf{GOT-10K.} The GOT-10k contains 10,000 high-quality video sequences, covering 580 object categories. The dataset includes over 1.5 million bounding box annotations in total. The dataset is evaluated using two metrics: Average Overlap (AO) and Success Rate (SR). As shown in \cref{tab:result}, our tracker demonstrated competitive performance across multiple metrics. MambaLCT-256 achieved 74.8\%, 85.4\%, and 72.1\% in metrics AO, \(SR_{0.5}\), and \(SR_{0.75}\), respectively. The high-resolution MambaLCT surpassed ARTrack-384, showing increases of 1.7\%, 2.4\%, and 2.9\% in terms AO, \(SR_{0.5}\), and \(SR_{0.75}\), respectively.

\textbf{TrackingNet.} The TrackingNet comprises over 30,000 video sequences, totaling more than 14 hours of video content. Each video sequence has an average length of 500 frames. The dataset particularly focuses on the stability and persistence of targets in long video sequences. 
As depicted in \cref{tab:result}, Our tracker achieved an AUC of 85.2\%, a P$_{\rm{Norm}}$ of 89.8\%, and a precision score of 83.9\%. The performance on the TrackingNet dataset clearly demonstrates the superiority of our tracker in long-term tracking.

\textbf{TNL2K.} TNL2K is a large-scale dataset for natural language tracking, containing approximately 2,000 video sequences, with a training and testing split ratio of 13:7. One notable feature of TNL2K is that it consists of multi-source data, including RGB-T, cartoons, and more. As illustrated in \cref{tab:tnl2k&uav}, MambaLCT exhibited competitive performance, achieving an AUC score of 58.5\%.
\begin{table*}[h]
    \centering
    % \caption{Comparison with state-of-the-art methods on TNL2K and UAV123 benchmarks in AUC score.}
    \resizebox{1\textwidth}{!}{
    \fontsize{8}{10}\selectfont
    \begin{tabular}{c|ccccccccccl|c}
     \toprule
     \multicolumn{1}{c|}{} & \multicolumn{1}{c}{SiamFC}& \multicolumn{1}{c}{MDNet} & \multicolumn{1}{c}{Ocean}
     & \multicolumn{1}{c}{TransT} & \multicolumn{1}{c}{TransInMo}&  ATOM &DiMP50 &STARK&\multicolumn{1}{c}{JointNLT}&\multicolumn{1}{c}{OSTrack} &SeqTrack& \multicolumn{1}{|c}{Ours} \\
     \midrule%第二道横线
     TNL2K& 29.5&38.0& 38.4& 50.7&52.0&  44.7 &44.7 & - &56.9 &55.9 &56.4& \textbf{58.5}\\
     UAV123& 46.8&52.8& 57.4& 68.1& 71.1&  64.3 &64.3 &68.2&-&68.3 &68.6& \textbf{70.1}\\
    \bottomrule 
     
    \end{tabular}
    }
    \caption{Comparison with state-of-the-art methods on TNL2K and UAV123 benchmarks in AUC score.}
    \label{tab:tnl2k&uav}
\end{table*}

\begin{table*}[h]
    \centering
    % \caption{Comparison with state-of-the-art methods on TNL2K and UAV123 benchmarks in AUC score.}
    \resizebox{1\textwidth}{!}{
    \fontsize{8}{10}\selectfont
    \begin{tabular}{ccccc|cccc|cccc}
     \toprule
     \multicolumn{5}{c|}{(a) Study on our method} & \multicolumn{4}{c|}{(b) Study on \(\boldsymbol{c}_p\) length} &\multicolumn{4}{c}{(c) Study on Mamba Layers} \\
     \midrule%
      \multicolumn{1}{c|}{Num} & \multicolumn{1}{c|}{Method} &\multicolumn{1}{c}{AUC} &\multicolumn{1}{c}{$\rm P_{norm}$} &\multicolumn{1}{c|}{P} &\multicolumn{1}{c|}{Sampling Length} &\multicolumn{1}{c}{AUC} &\multicolumn{1}{c}{$\rm P_{norm}$} &\multicolumn{1}{c|}{P} &\multicolumn{1}{c|}{Mamba Layers} &\multicolumn{1}{c}{AUC} &\multicolumn{1}{c}{$\rm P_{norm}$} &\multicolumn{1}{c}{P} \\
      \midrule%
      \multicolumn{1}{c|}{1} &\multicolumn{1}{c|}{baseline} &69.1\%& 78.7\% &\multicolumn{1}{c|}{75.2\%} &\multicolumn{1}{c|}{1} & \textbf{71.8\%} & \textbf{83.0\%} &\multicolumn{1}{c|}{\textbf{79.4\%}} &\multicolumn{1}{c|}{None} & 70.1\% & 80.2\% &78.6\%\\
      
      \multicolumn{1}{c|}{2} &\multicolumn{1}{c|}{+HiViT} &70.5\%& 80.7\% &\multicolumn{1}{c|}{77.5\%} &\multicolumn{1}{c|}{2} & 71.3\%& 82.7\%&\multicolumn{1}{c|}{79.1\%} &\multicolumn{1}{c|}{(1,2,3)} & 70.5\%& 81.9\%&77.5\%\\
      
      \multicolumn{1}{c|}{3} &\multicolumn{1}{c|}{+sequen sampling} & 70.1\% & 80.2\% &\multicolumn{1}{c|}{78.6\%} &\multicolumn{1}{c|}{3} &70.8\%& 82.1\% &\multicolumn{1}{c|}{78.0\%} &\multicolumn{1}{c|}{(3,6,9)} &\textbf{71.8\%} & \textbf{83.0\%} &\textbf{79.4\%}\\
      
      \multicolumn{1}{c|}{4} &\multicolumn{1}{c|}{+Mamba} &\textbf{71.8\%} & \textbf{83.0\%} &\multicolumn{1}{c|}{\textbf{79.4\%}} &\multicolumn{1}{c|}{4} & 69.8\%& 81.0\%&\multicolumn{1}{c|}{76.7\%} &\multicolumn{1}{c|}{(18,19,20)} & 70.9\%& 82.1\%&78.0\%\\
    \bottomrule% 
     
    \end{tabular}
    }
    \caption{Three ablation studies on the LaSOT benchmark.}
    \label{tab:ablation study}
\end{table*}

\textbf{UAV123.} UAV123 is a dataset for low-altitude UAV object tracking, comprising a total of 123 video sequences. As evidenced by the \cref{tab:tnl2k&uav}, compared to SeqTrack and OStrack, MambaLCT achieved improvements of 1.5\% and 1.8\% in AUC scores, respectively.

\subsection{Ablation Study}

To explore the effectiveness of our proposed method, we conducted extensive experiments using the MambaLCT-256 model on the LaSOT dataset.

\textbf{Baseline Vs MambaLCT.} Our MambaLCT is based on the OSTrack framework. Specifically, we replaced ViT with HiViT, used sequence sampling instead of single-frame sampling, and enhanced object perception within the background by learning video context information through Mamba. In \cref{tab:ablation study}(a), \#1 indicates that the hierarchical ViT significantly improves our model's ability to extract spatial information, with a 0.6\% increase in the AUC score. According to the results of \#2 and \#3, we can observe that using video sampling alone does not improve the performance of the tracker. However, as shown in \#4, effectively utilizing context information from video sampling can significantly enhance the performance of the tracker, achieved 71.8\%, 83.0\%, and 79.4\% in AUC, P$_{\rm{Norm}}$, and precision score.

\textbf{Study on length of \(\boldsymbol{c}_p\).} 
To verify the impact of context information token length on experimental performance, we designed a series of experiments using different lengths of \(\boldsymbol{c}_p\). As shown in \cref{tab:ablation study}(b), when the length of CP is 1, the tracker achieved 71.8\%, 83.0\%, and 79.4\% on the three metrics, respectively. With the length increases, the performance gradually decreases. We believe that this phenomenon is due to the autoregressive nature of Mamba, where each token's generation depends on the preceding tokens. \(\boldsymbol{c}_p\) composed of too many consecutive tokens can lead to redundant contextual information.

\textbf{Study on different Mamba layers.}
We evaluated the different insertion layers of the Mamba module we used and summarized the results in \cref{tab:ablation study}(c). Compared to inserting Mamba into the first three and last three layers of the backbone network, the method of uniformly inserting Mamba throughout the network can learn multi-level contextual information, achieving a maximum AUC score of 71.8\%.

% \begin{table}[t]
%     \centering
%     \caption{Ablation studies for our method }
%     \fontsize{9}{10.8}\selectfont
%     \begin{tabular}{c|l|ccc}
%      \toprule
%      \multicolumn{1}{c|}{Num} & \multicolumn{1}{l|}{Method} & \multicolumn{1}{c}{AUC}
%      & \multicolumn{1}{c}{$\rm P_{norm}$} & \multicolumn{1}{c}{P} \\
%      \midrule%第二道横线
%      1& baseline& 69.1\%& 78.7\% &75.2\% \\
%      2& +HiViT& 70.5\%& 80.7\% & 77.5\% \\
%      3& +sequen sampling& 70.1\% & 80.2\% &78.6\% \\
%      4& +Mamba& 71.8\% & 83.0\% &79.4\% \\
%      \bottomrule

%     \end{tabular}

%     \label{tab:study1}
% \end{table}

\begin{figure}[t]
      \centering
       \includegraphics[width=1\linewidth]{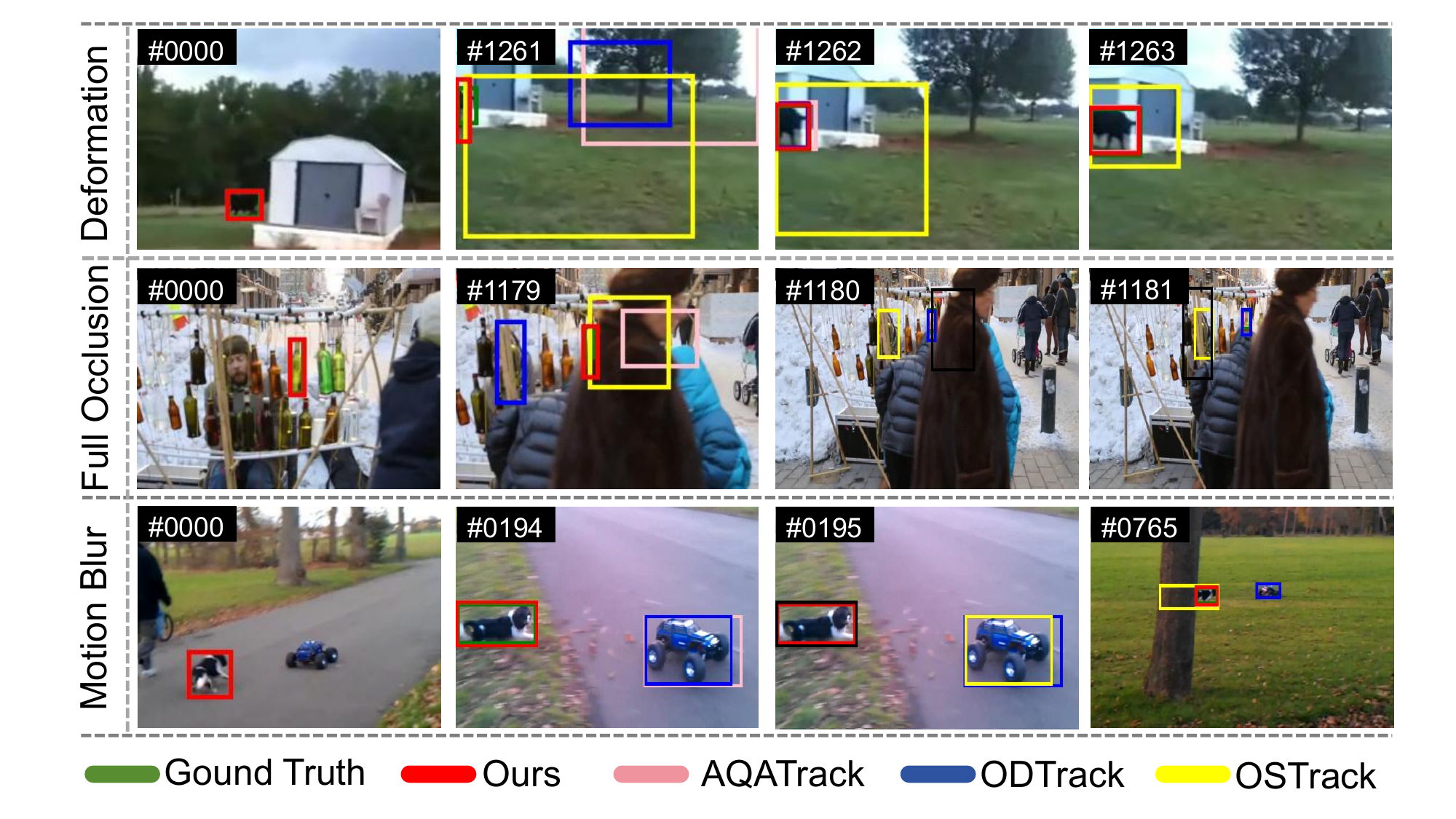}
    
       \caption{On the LaSOT benchmark, we visualized the comparison results of our tracker with three SOTA trackers across three challenges. }
       \label{figure:bbx}
       \vskip -0.1in
\end{figure}

\subsection{Visualization and Limitation}
\textbf{Visualization.} To visually demonstrate the effectiveness of our proposed method, we visualized the AUC score of MambaLCT under different challenges on the LaSOT benchmark, as shown in \cref{fig:eao_lasot}. Compared to other video-level trackers, MambaLCT excels in challenges such as motion blur, full occlusion, and deformation. As illustrated in \cref{figure:bbx}, we visualized the tracking results of sequences under three challenges.
From these results, we can see that thanks to the increased length of context information, our tracker can more accurately locate the target in complex situations such as target deformation and occlusion, compared to other trackers.

\begin{figure}[t]
      \centering
       \includegraphics[width=1\linewidth]{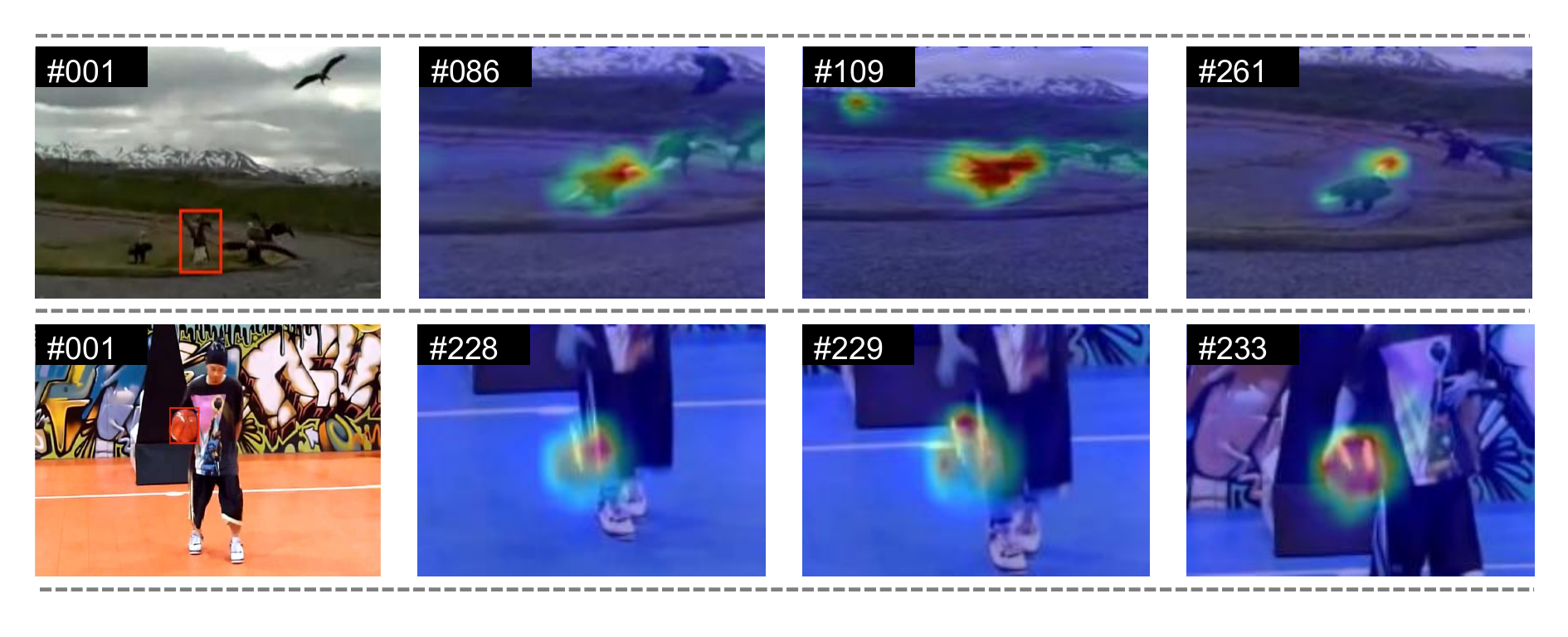}
    
       \caption{Visualization of the attention to the search frame by the cross-frame context information. As the length of the context information increases, the attention to the target becomes more focused. }
       \label{figure:att}
       % \vskip -0.1in
\end{figure}

Additionally, to verify whether the context information can capture the target's information, we visualized the attention maps of the context information on the search frame, as shown in \cref{figure:att}. We found that as the sequence length increases, the attention of the context information becomes more focused on the target, which also demonstrates the advantages of long-term context information.

\textbf{Limitation.} Our proposed video-level context information modeling method effectively captures the complete behavior and contextual understanding of the target, thereby enhancing the model's perception of the target. Despite achieving significant results, our proposed method faces limitations due to computational resource constraints. Consequently, the training and testing phases cannot be unified. During the training phase, we do not model the entire video sequence but instead sample portions of the sequence. Designing a more coherent training strategy is a direction for our future work. One feasible direction is to replace the previous image sampling with sequence sampling.

\section{Conclusion}
In this paper, we introduced MambaLCT, a novel tracking framework designed to enhance object perception by constructing long-term context information. We extended the context information length from the initial frame to the current frame, which provides a more comprehensive overview of the target's historical information. We combine the strengths of Mamba and Transformers to construct and utilize long-term contextual information. The Transformer learns the appearance features of the images in an autoregressive manner, while Mamba extracts long-term contextual information about the target from historical appearance data. Detailed experiments show that our MambaLCT achieves excellent results across six tracking benchmarks, although our work still has limitations. We hope this work can contribute valuable insights and assistance to current tracking paradigms.

\section{Acknowledgments}
This work is supported by the Project of Guangxi Science and Technology (No.2024GXNSFGA010001 and 2022GXNSFDA035079), the National Natural Science Foundation of China (No.U23A20383 and 62472109), the Guangxi ”Young Bagui Scholar” Teams for Innovation and Research Project, the Research Project of Guangxi Normal University (No.2024DF001), the Innovation Project of Guangxi Graduate Education (YCBZ2024083), and the grant from Guangxi Colleges and Universities Key Laboratory of Intelligent Software (No.2024B01).

\bibliography{aaai25}

\end{document}